\documentclass[11pt,a4paper]{article}
\usepackage[hyperref]{emnlp2020}
\usepackage{times}
\usepackage{latexsym}
\usepackage{graphicx}
\usepackage{amsmath,empheq}
\usepackage{subcaption}
\usepackage{mdframed}
\usepackage{multirow}
\usepackage{url}
\usepackage{booktabs,tabularx,enumitem,ragged2e}

\usepackage{ulem}
\usepackage{microtype}

\aclfinalcopy 

\newcommand{\model}{\textsc{PRover}}

\title{\textsc{PRover}: Proof Generation for Interpretable Reasoning over Rules}

\author{Swarnadeep Saha \;\;\;\;\;\; Sayan Ghosh \;\;\;\;\;\; Shashank Srivastava \;\;\;\;\;\; Mohit Bansal
\\ 
  UNC Chapel Hill\\ 
  \texttt{\{swarna, sayghosh, ssrivastava, mbansal\}@cs.unc.edu}
}

\date{}

\begin{document}
\maketitle
\begin{abstract}
Recent work by \citet{clark2020transformers} shows that transformers can act as ``soft theorem provers'' by answering questions over explicitly provided knowledge in natural language. In our work, we take a step closer to emulating formal theorem provers, by proposing \textsc{PRover}, an interpretable transformer-based model that jointly answers binary questions over rule-bases and generates the corresponding proofs. Our model learns to predict nodes and edges corresponding to proof graphs in an efficient constrained training paradigm. During inference, a valid proof, satisfying a set of global constraints is generated. We conduct experiments on synthetic, hand-authored, and human-paraphrased rule-bases to show promising results for QA and proof generation, with strong generalization performance. First, \textsc{PRover} generates proofs with an accuracy of 87\%, while retaining or improving performance on the QA task, compared to RuleTakers  (up to 6\% improvement on zero-shot evaluation). Second, when trained on questions requiring lower depths of reasoning, it generalizes significantly better to higher depths (up to 15\% improvement). Third, \textsc{PRover} obtains near perfect QA accuracy of 98\% using only 40\% of the training data.
However, generating proofs for questions requiring higher depths of reasoning becomes challenging, and the accuracy drops to 65\% for ``depth 5", indicating significant scope for future work.\footnote{Our code and models are publicly available at \url{https://github.com/swarnaHub/PRover}.}
\end{abstract}

\begin{figure}
    \centering
    \includegraphics[width=0.8\columnwidth]{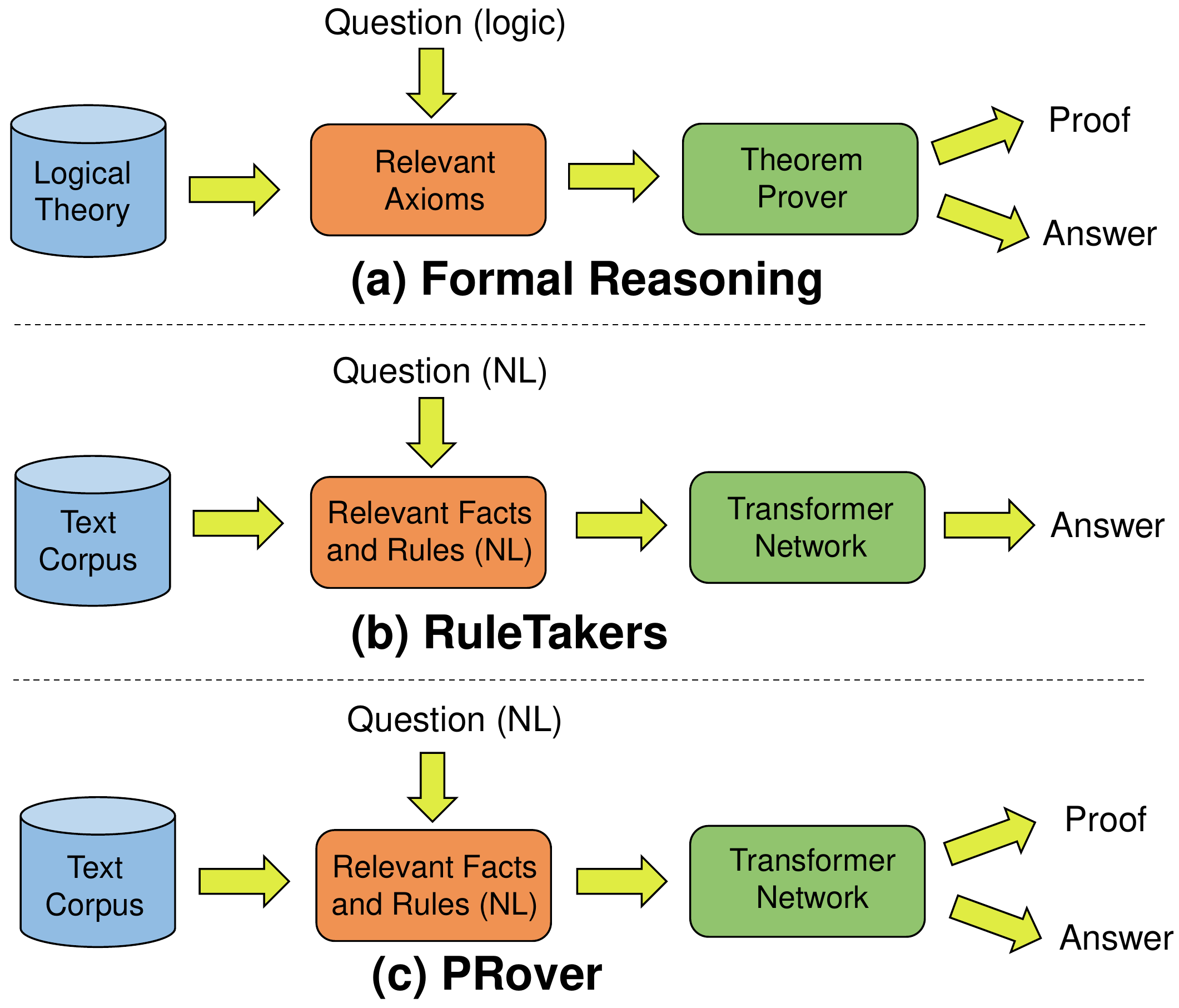}
    \caption{Block diagram showing that \textsc{PRover} is a closer linguistic analog of formal reasoning.
    \vspace{-3pt}}
    \vspace{-10pt}
    \label{fig:top_figure}
\end{figure}

\section{Introduction}

\begin{figure*}
\centering
    \includegraphics[width=0.85\textwidth]{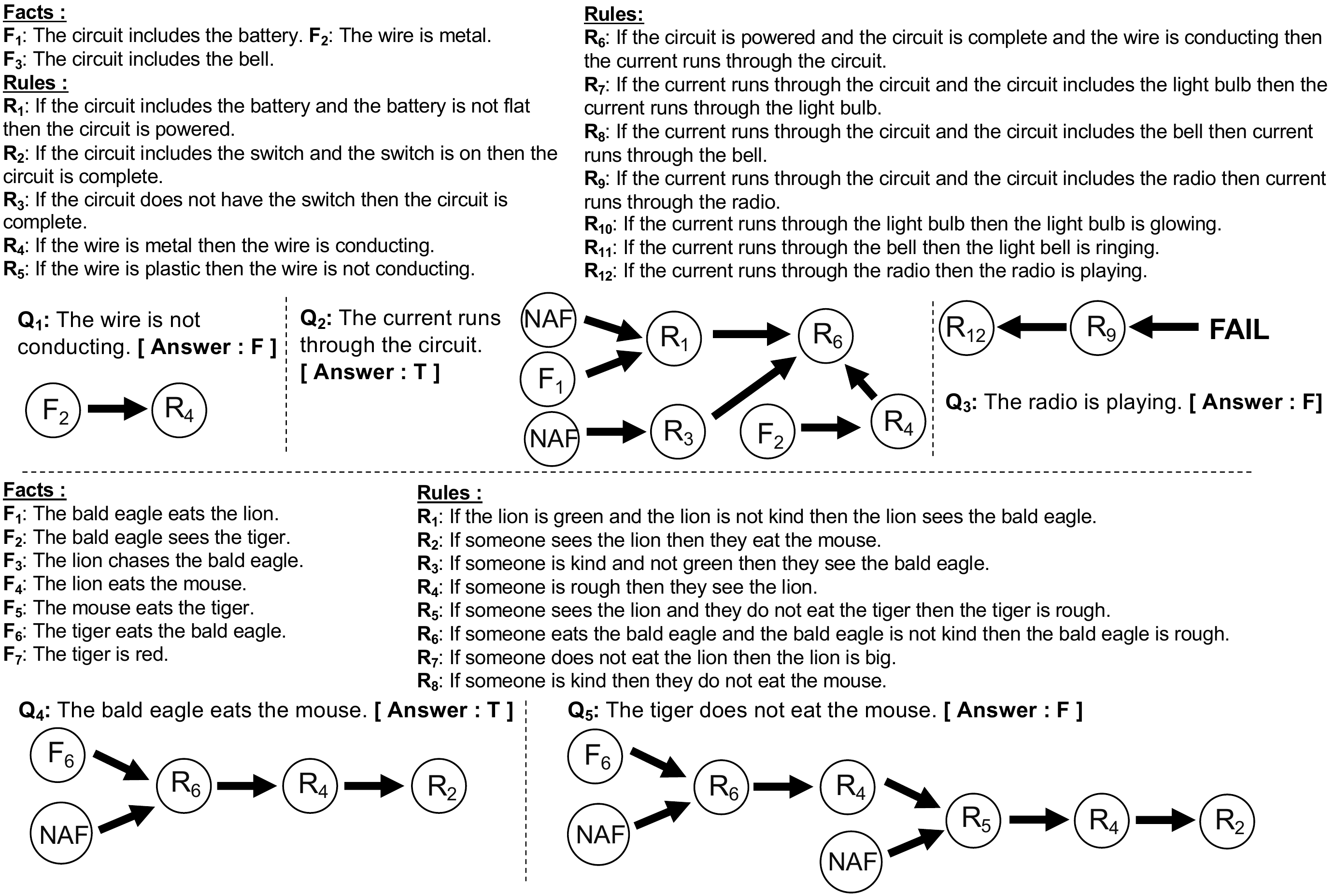}
    \vspace{-5pt}
\caption{Diagram showing two rule-bases with rules, facts, questions, answers and proofs. \textsc{PRover} answers all the questions correctly and also generates all the corresponding proofs accurately in the above scenarios. \vspace{-5pt}
\vspace{-5pt}}
\label{fig:example_dataset_proof}
\end{figure*}

Developing systems that can understand and reason over explicitly provided knowledge has been a fundamental goal of AI~\cite{newell1956logic}. Owing to the challenges posed in reasoning over formal representations~\cite{musen1988brittleness}, and backed by the recent successes of transformers \cite{vaswani2017attention} in NLP, ~\citet{clark2020transformers} propose a new version of the problem by replacing the formal representations of rule-bases with natural language (English). Specifically, their task requires predicting the truth value of a statement by reasoning over a set of facts and rules, all expressed in natural language. Figure \ref{fig:example_dataset_proof} shows some examples of the task. \citet{clark2020transformers} propose RuleTakers, a fine-tuned RoBERTa model \cite{liu2019roberta} to show that transformers can act as ``soft theorem provers'' by predicting the final answer in such reasoning-based problems with high accuracy. 

We argue that to use transformers for natural language reasoning reliably, they should be able to generate proofs that provide rationales for the predicted answer. Proof generation is vital for emulating formal reasoning but also for moving towards human-interpretable models that alleviate concerns about the black-box nature of deep architectures \cite{rudin2019stop}.
Towards this, we present \textsc{PRover}, a transformer-based model that jointly answers questions over natural language rule-bases and generates corresponding proofs. Figure \ref{fig:top_figure} illustrates our method as a closer linguistic analog of formal reasoning, as it generates proofs along with answers. However, unlike formal reasoners, \textsc{PRover} can operate on natural language text that provides the underlying theory, rather than rely on formal logical representations. Such methods that combine interpretability and flexibility in reasoning can have wide applications across domains.

\textsc{PRover}'s architecture consists of three modules that together generate answers along with proofs. In this work, proofs are represented as directed graphs consisting of the relevant rules and facts needed to prove or disprove the question statement. Section \ref{proof-section} contains details of this representation. A QA module predicts a binary answer for the question, a node module chooses which rules and facts are part of the proof, and an edge module predicts the presence and the direction of the edges between the chosen nodes. Model training minimizes a joint cross-entropy loss over the three modules. To guide the model to predict edges between valid nodes only, we enforce global constraints on the structure of the proof during training, by masking out labels for impossible edges, resulting in a more efficient learning problem. \textsc{PRover} generates valid proofs during inference by solving an ILP over the edge potentials, subject to multiple semantic constraints, such as ensuring proof graph connectivity. Our contributions are:

\begin{itemize}[nosep, wide=0pt, leftmargin=*, after=\strut]
    \item We present \textsc{PRover}, an interpretable joint model that learns to reason over natural language rule-bases and generate corresponding proofs.
    \item \textsc{PRover} performs similarly or improves upon state-of-the-art QA accuracy for the task, with up to 6\% improvement on zero-shot evaluation, and generates exact proofs at 87\% accuracy. Unlike RuleTakers, it does not require additional fine-tuning on the RACE \cite{lai2017race} dataset.
    \item \textsc{PRover} demonstrates significantly better generalization. When trained on lower depth questions, it shows better QA accuracy (up to 15\%) on higher depth ones.
\end{itemize}{}

\section{Related Work}
Our work is related to multiple bodies of previous work in NLP and formal reasoning.

\paragraph{QA and NLI:}
The rule reasoning task is related to reasoning tasks that have been proposed recently. These include tasks in the bAbI dataset \cite{weston2015towards}, synthetically generated probe tasks ~\cite{richardson2019probing} or reading comprehension tasks in datasets such as QuaRTz \cite{tafjord-etal-2019-quartz} and ROPES \cite{lin-etal-2019-reasoning}. Unlike our task, most of these require reasoning over implicit rules, the focus being on language understanding and one step of rule application. Multi-hop QA datasets like HotpotQA \cite{yang2018hotpotqa} require multiple reasoning steps, but the inference rules needed are again implicitly inferred, rather than explicitly provided. Our task also bears similarity with Natural Language Inference \cite{maccartney2014natural}, but NLI also allows unsupported inferences by filling gaps in explicitly stated knowledge \cite{dagan2013recognizing}.

\paragraph{Formal Reasoning and Neural Theorem Proving:}
Semantic parsing~\cite{zettlemoyer2012learning, berant2013semantic, berant2014semantic} of multi-sentence texts into logical forms has proved to be challenging, restricting the application of semantic parsers to formal reasoning systems \cite{kamath2018survey}. \textsc{PRover} bypasses this expensive and error-prone process and attempts to solve the problem in an end-to-end manner, without any intermediate logical representations.

Our approach is conceptually similar to a body of work on Neural Theorem Proving \cite{weber-etal-2019-nlprolog} that has focused on developing theorem provers by combining reasoning from symbolic techniques with the possibility of differentiable learning from neural networks. These include neuro-symbolic methods for table comprehension \cite{neelakantan2015neural}, executing basic compositional programs~\cite{reed2015neural}, SAT solving~\cite{selsam2018learning}, formula embedding~\cite{abdelaziz2020experimental}, approximate (DNF) model counting~\cite{abboud2019learning}, etc. However, \textsc{PRover} diverges from these in working with free-form natural language input to generate proofs similar to formal reasoners.

\paragraph{Model Interpretability:}
\textsc{PRover} follows a significant body of previous work on developing interpretable neural models for NLP tasks to foster explainability. Several approaches have focused on formalizing the notion of interpretability \cite{rudin2019stop, doshi2017towards, hase_evaluating_2020}, tweaking features for local model interpretability \cite{ribeiro2016should, ribeiro2018anchors} and exploring interpretability in latent spaces \cite{joshi2018xgems, samangouei2018explaingan}. Our work can be seen as generating explanations in the form of proofs for an NLP task. While there has been prior work on generating natural language explanations for multiple NLP tasks, including NLI \cite{camburu2018snli}, commonsense reasoning \cite{rajani2019explain, zhang2020WinoWhy} and generic text classification tasks \cite{liu2019towards}, our novelty lies in generating compositional explanations consisting of proof graphs that detail the chain of reasoning, starting from language. We use a max-flow ILP formulation for checking proof graph connectivity \cite{even1975network}. Multiple approaches for NLP tasks such as sentiment analysis and content selection \cite{pang2004sentimental, barzilay2005collective,  bansal2008power} have been framed as optimal flow problems on graphs.

\paragraph{Program Synthesis with Transformers:} 
Existing works show that transformers already capture some knowledge from pre-training for algorithm emulation \cite{talmor2019olmpics} or can be fine-tuned for tasks like semantic parsing \cite{he2020establishing}, translation \cite{wang-etal-2019-learning}, symbolic integration \cite{Lample2020Deep} and mathematics \cite{saxton2018analysing}. In our work, we also employ a transformer-based pre-trained language model (RoBERTa \cite{liu2019roberta}) but for the downstream task of rule-based reasoning. 

\section{Method}
Each input to \textsc{PRover} is a context $C$ (consisting of facts $F$ and rules $R$) and a question $Q$, about the context. \textsc{PRover} predicts the answer $A \in \{True, False\}$ and generates a proof $\mathcal{P}$.

\subsection{Proof Representation}
\label{proof-section}
A proof, $\mathcal{P} = ( \mathcal{N}, \mathcal{E})$, is a directed graph with nodes $n \in \mathcal{N}$ and edges $e \in \mathcal{E}$. Each node is either a fact $f$ $\in$ $F$, a rule $r$ $\in$ $R$ or a special $\mathit{NAF}$ node (Negation As Failure, as described below). Edges in the proof are directed either from a fact (or $\mathit{NAF}$) to a rule or from a rule to another rule. These indicate that a fact (or $\mathit{NAF}$) is consumed by a rule, or the output of a rule (a new fact) is consumed by another rule, respectively. We use these constraints both during \textsc{PRover}'s training and inference, as described later in the paper. Formally, we have:
\begin{eqnarray}
    \mathcal{P} = (\mathcal{N},\mathcal{E}) \nonumber \\
    \mathcal{N} \subseteq R\cup F \cup \mathit{NAF} \nonumber \\
    \mathcal{E} \subseteq \mathcal{N} \times \mathcal{N} \nonumber
\end{eqnarray}
Figure \ref{fig:example_dataset_proof} shows examples of two contexts (consisting of facts and rules), five questions about the contexts, along with their answers and proofs.
Each proof has a depth ($Q_1$'s proof has a depth of 1). The maximum proof depth in all the datasets considered in this work \cite{clark2020transformers} is 5. 
Proofs in the datasets are of three types:

\paragraph{\textbf{Successful proof with NAF}:} The proof of $Q_1$ in Figure \ref{fig:example_dataset_proof} is one such such example. $F_2$ acts on $R_4$ to prove that ``The wire is conducting.'' and hence the answer is false.
\paragraph{\textbf{Successful proof with NAF}:} Given a statement $s$, $\mathit{NAF}$ in logic programming is a non-monotonic inference rule used to derive ``not $s$'' (negation of the statement) from failure to derive $s$. Hence, a proof may contain $\mathit{NAF}$ node(s), representing the truthfulness of the negation of statement(s) that cannot be proved using the set of rules. Consider the proofs for $Q_4$ and $Q_5$ where the $\mathit{NAF}$ node in $Q_4$ represents ``The bald eagle is not kind.''.
\paragraph{\textbf{Failed proof}:} This happens when a statement cannot be derived using the given rule-base and the shallowest branch of the proof tree that fails is shown. $Q_3$'s proof in Figure \ref{fig:example_dataset_proof} is an example as ``The radio is playing.'' cannot be proved.

Note that a proof can have edges between two rules in both directions. E.g., consider the edges $R_4 \rightarrow R_5$ and $R_5 \rightarrow R_4$ in $Q_5$'s proof in Figure~\ref{fig:example_dataset_proof}. A node can have more than two incoming edges -- the node $R_6$ in $Q_2$ has three incoming edges from $R_1$, $R_3$, and $R_4$.

\subsection{Task Description}
\label{const_train}
Each training example is a tuple $(C_i:=\{F_i, R_i\}, Q_i, A_i, \mathcal{P}_i )$ consisting of a context (set of rules and facts), a question, the corresponding answer, and a proof. Generating a proof graph requires (1) identifying the nodes (set of relevant facts, $\mathit{NAF}$ and rules) that are part of the proof, (2) identifying the edges connecting these nodes, and (3) verifying a set of global constraints such as proof connectivity that ensure a valid proof. 

For the first, we predict a binary label over each rule, fact and $\mathit{NAF}$ denoting their presence or absence in the proof. For the second, we also predict binary labels denoting the presence or absence of each edge. For the third, we enforce constraints during both training and inference (Section \ref{ilp}). During training, we mask out the edge labels\footnote{The masked edges do not contribute to the training loss.} corresponding to (1) self-loops, (2) edges between absent nodes, and (3) edges between facts to facts and rules to facts. This enforces a semantic constraint that the set of candidate edges in the ensuing proof is consistent with the chosen set of nodes, and also simplifies the learning problem, since a smaller number of edges need to be labeled.

\begin{figure*}
    \centering
    \includegraphics[width=0.9\textwidth]{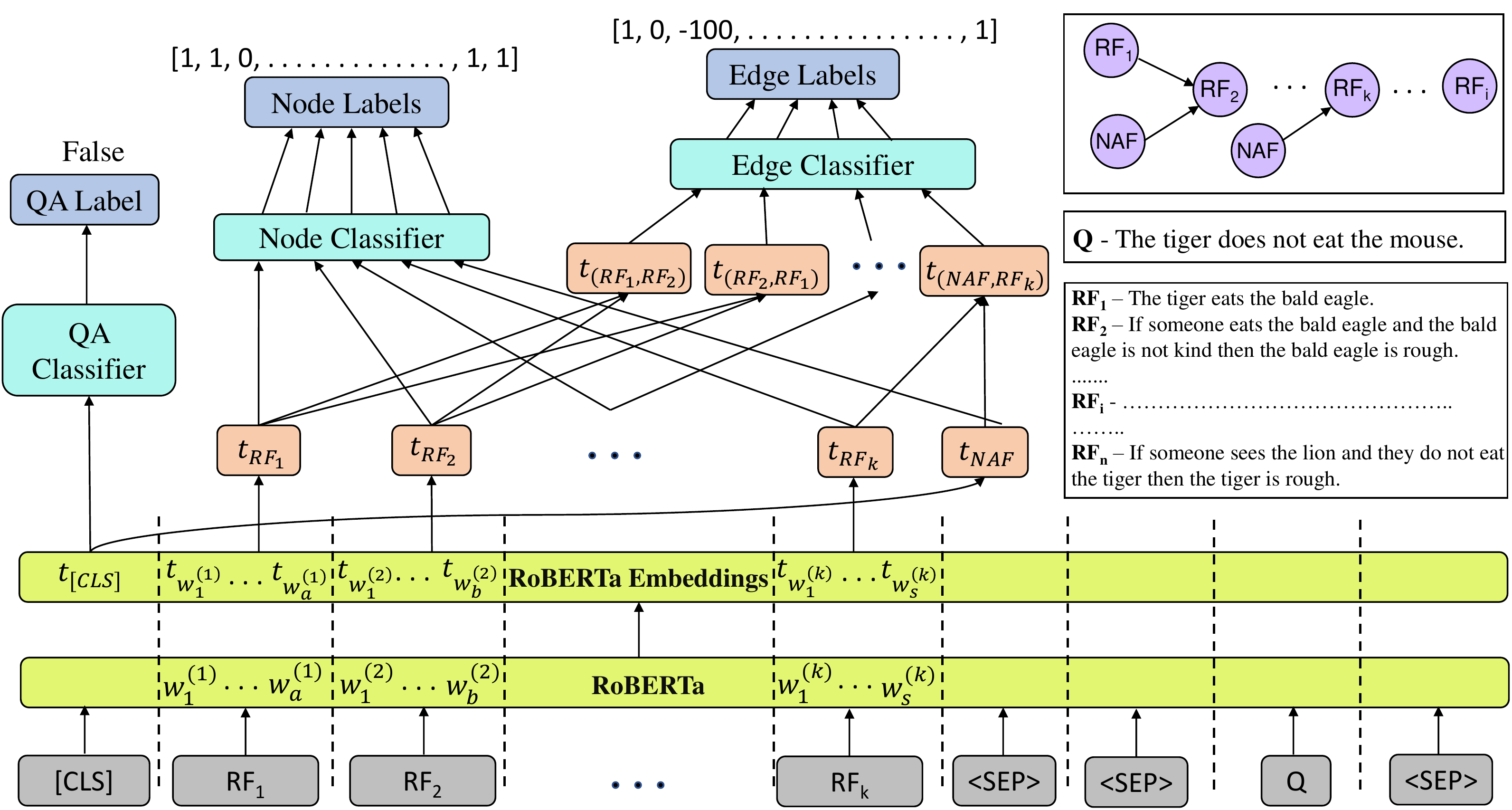}
    \vspace{-5pt}
    \caption{\small Architecture diagram of \textsc{PRover}. The presence and absence of nodes/edges are labeled by 1 and 0 respectively while -100 represents masked out edges.
    \vspace{-12pt}}
    \label{fig:model}
\end{figure*}{}

\subsection{\textsc{PRover}: Joint QA and Proof Generation Model}
\label{model}
Figure \ref{fig:model} shows the architecture of \textsc{PRover}, built on top of RoBERTa \cite{liu2019roberta}. Our model consists of three modules: (1) QA module, (2) Node module, and (3) Edge module. The QA module is exactly the same as the RuleTakers model \cite{clark2020transformers}, thus allowing us to directly evaluate the effectiveness of our node and edge modules.  The input to RoBERTa is the concatenation of the context and the question, separated by the $[SEP]$ tokens. The context is represented by concatenating the text consisting of facts and rules. Formally, if the rules and facts are denoted by $\{RF_i\}_{i=1}^k$ and the question by $Q$, the input is
\begin{equation}
    [CLS]\hspace{1ex} \{RF_i\}_{i=1}^k \hspace{1ex} [SEP] [SEP] \hspace{1ex} Q \hspace{1ex} [SEP] \nonumber
\end{equation}

\paragraph{QA Module:} The output of RoBERTa contains an embedding for each token in the context and a global embedding corresponding to the $[CLS]$ token. The QA classification head $H_{QA}$ is a sequence of two linear layers with dropout probability of $p$. Formally, if $t_{[CLS]}$ denotes the $[CLS]$ token embedding, we obtain the class-wise probability values $P_{QA}$ using the softmax function $\sigma$.
\begin{equation}
    P_{QA} = \sigma(H_{QA}(t_{[CLS]})) \nonumber
\end{equation}
\paragraph{Node Module:} Let  $\{w_j^{(i)}\}_{j=1}^{m}$ denotes the $m$ tokens corresponding to $RF_i$. Assuming the corresponding RoBERTa embeddings are denoted by $\{t_{w_j^{(i)}}\}_{j=1}^{m}$, we learn a representation $t_{RF_i}$ for each $RF_i$, by performing a mean pooling $\mathit{MP}$ of the constituent token embeddings.
\begin{equation}
    t_{RF_i} = \mathit{MP}(\{t_{w_j^{(i)} }\}_{j=1}^{m}) \nonumber
\end{equation}
\noindent We also learn a representation of the $\mathit{NAF}$ node $t_{NAF}$ as a linear transformation on $t_{[CLS]}$. Note that due to the self-attention layers of RoBERTa, $t_{[CLS]}$ summarizes the set of all derivable facts given the context and the question. We want the $\mathit{NAF}$ node to encode information about all facts containing negation (e.g.,``The bald eagle is not kind'' in $Q_4$'s proof of Figure \ref{fig:example_dataset_proof}) in the context. These are taken as true as their positive counterparts (``The bald eagle is kind'') are non-derivable given the context. Thus, if a statement $s$ cannot be derived from the facts and the rules in a context, the $\mathit{NAF}$ node should infer that ``not $s$'' is true. We model this notion of the negation of all unprovable statements (given a context) by learning $\mathit{NAF}$ as a function of everything provable in the context, encoded by the $t_{[CLS]}$ embedding.\footnote{We note that a proof can have multiple $\mathit{NAF}$ nodes, each representing a different negated fact. Since the datasets label all these as $\mathit{NAF}$, we collapse all the $\mathit{NAF}$ nodes into a single node and learn a unified representation for them.}

\noindent The node classifier $H_{Node}$ has a similar architecture to the QA classifier and predicts a presence and absence probability score for each node.
\begin{equation}
    P_{Node} = \sigma(H_{Node}(\{t_{RF_i}\}_{i=1}^k, t_{\mathit{NAF}})) \nonumber
\end{equation}
\paragraph{Edge Module:} Now, given the representations of each fact, rule and $\mathit{NAF}$, we learn a representation for each edge between these. Formally, we define the edge embedding $t_{(RF_i, RF_j)}$ from node $RF_i$ to node $RF_j$ by concatenating their individual embeddings $t_{RF_i}$ and $t_{RF_j}$ with their element-wise difference (which gives the directionality vector). 

\begin{equation}
    t_{(RF_i, RF_j)} = [t_{RF_i}, t_{RF_j}, (t_{RF_j}-t_{RF_i})] \nonumber
\end{equation}
\noindent The above formulation also helps learn separate representations for edges $RF_i \rightarrow RF_j$ and $RF_j \rightarrow RF_i$. This is essential for our task as a proof can have edges between two rules in both directions. In Section~\ref{ablate}, we see that this formulation leads to a near perfect empirical performance in predicting the directionality of edges. The edge classifier $H_{Edge}$ outputs probability scores representing the presence and absence of each edge.
\begin{eqnarray}
    P_{Edge} = \sigma(H_{Edge}(\{{t_{(RF_i,RF_j)}\}_{i,j=1}^{k+1}})) \nonumber \\
    t_{RF_{k+1}} = t_{\mathit{NAF}} \nonumber
\end{eqnarray}{}
We train our model by using binary cross-entropy loss for each of the three modules. Formally, if $L_{QA}$, $L_{Node}$ and $L_{Edge}$ denote the three losses, the overall loss $L$ is given by:
\begin{equation}
    L = L_{QA} + L_{Node} + L_{Edge} \nonumber
\end{equation}

\subsection{ILP Inference for Global Constraints}
\label{ilp}
As mentioned previously, during inference, we enforce additional constraints on the structure of the predicted proof graph. For this, we frame inference as Integer Linear Program (ILP) optimization, which we describe next. We follow the generative process of a graph wherein the nodes are defined first, followed by the edges on that set of nodes. Thus, we fix the nodes first based on the predictions by the node module of our model and maximize a global score over the set of edges only. This reduces the large search space and ensures that all constraints can be expressed as linear expressions.

\paragraph{Proof Connectivity Formulation:} An important constraint is to ensure that the predicted proof graphs are connected.\footnote{Proofs are directed graphs. We check connectivity in the equivalent undirected graphs.} To check if a proof graph $\mathcal{P}$ is connected, we define an augmented graph $\mathcal{P}_{aug} = (\mathcal{N}_{aug}, \mathcal{E}_{aug})$ with two added nodes ``source" and ``sink". We add an edge from the source to any one of the nodes $x$ in $\mathcal{P}$. We also define edges from all nodes in $\mathcal{P}$ to the sink.
\begin{eqnarray}
\mathcal{N}_{aug} = \mathcal{N} \cup \{source, sink\} \nonumber \\
\mathcal{E}_{aug} = \mathcal{E} \cup \{(source, x)\} \cup \{(n, sink) \forall n \in \mathcal{N}\} \nonumber
\end{eqnarray}

\noindent Having defined $\mathcal{P}_{aug}$, we can reduce the graph connectivity in $\mathcal{P}$ to a maximum flow problem \cite{leighton1999multicommodity} in $\mathcal{P}_{aug}$ \cite{even1975network}. For this, we define the capacity variable $c_{(m,n)}$ for each edge, $m \rightarrow n$ in $P_{aug}$, as follows.
\begin{eqnarray}
c_{(source,x)} = |\mathcal{N}| \text{ and } c_{(x, source)} = 0 \nonumber \\
\forall n \in \mathcal{N}, c_{(n,sink)} = 1 \text{ and } c_{(sink,n)} = 0 \nonumber \\
\forall m,n \in \mathcal{N},  c_{(m,n)} = |\mathcal{N}| \nonumber \\
c_{(m,n)} = 0 \text{ if } m \notin \mathcal{N} \text{ or } n \notin \mathcal{N} \nonumber 
\end{eqnarray}

\noindent Now, there can be a maximum total flow of $|\mathcal{N}|$ from the source to the sink, if and only if the graph is connected. We use this flow formulation to provide additional constraints for our ILP inference procedure that ensure connectivity of proof graphs.

\paragraph{\textbf{Final Optimization Problem:}} Our maximization objective, subject to the connectivity constraint and all other constraints (that ensure a valid proof) is as follows. Let $\phi_{(m,n)}$ represent the probability that an edge $m \rightarrow n$ is present, as predicted by \textsc{PRover}. We want to infer 0/1 assignments for our optimization variables $e_{(m,n)}$ (a value of 1 means the edge is part of the proof, while 0 means it is not) such that the following objective is maximized:
\begin{equation}
\begin{split}
    \underset{e_{(m,n)}, f_{(m,n)}}{argmax} \sum_{m,n, m\neq n} (\phi_{(m,n)}e_{(m,n)} + \\ (1-\phi_{(m,n)})(1-e_{(m,n)})) \nonumber
\end{split}
\end{equation}
subject to constraints: 
\begin{eqnarray}
\small
    \forall m,n \in F\cup R \cup \mathit{NAF},  e_{(m,n)} \in \{0,1\}\label{eq:const1} \\
    e_{(m,n)} = 0, \text{if } m \notin \mathcal{N} \text{or } n \notin \mathcal{N} \label{eq:const2} \\
    e_{(m,n)} = 0, \text{if } m \in F \text{ and } n \in F \label{eq:const3} \\
    e_{(m,n)} = 0, \text{if } m \in R \text{ and } n \in F \label{eq:const4} \\
    \forall m,n \in \mathcal{N}_{aug}, 0 \leq f_{(m,n)} \leq c_{(m,n)} \label{eq:const5}\\
    \forall n \in \mathcal{N}_{aug}, \sum_{m:(m,n)\in \mathcal{E}_{aug}} f_{(m,n)} \nonumber \\ =  \sum_{o:(n,o)\in \mathcal{E}_{aug}} f_{(n,o)}\label{eq:const7}\\
    f_{(source, x)} = |\mathcal{N}| \label{eq:const8}\\
    \forall m,n \in \mathcal{N}_{aug}, e_{(m,n)} + e_{(n,m)} \nonumber \\ - (f_{(m,n)} / |\mathcal{N}|) \geq 0 \label{eq:const9}
\end{eqnarray}

Note that $\mathcal{N}$, $F$, and $R$ refer to the set of predicted nodes (from the model), the set of facts, and the set of rules, respectively. Equations \ref{eq:const2}, \ref{eq:const3} and \ref{eq:const4} ensure that edges are present only when the corresponding nodes are present and that there are no edges between two facts and from a rule to a fact. Next, to ensure proof connectivity, we first define the flow constraints in Equations \ref{eq:const5} and \ref{eq:const7} constrained by the flow variables $f_{(m,n)}$ for each edge $m \rightarrow n$. These maintain the capacity constraints (the flow at each edge should be less than its capacity) and the flow conservation constraints (the total flow through the incoming edges at a node is equal to the total flow through the outgoing edges). Equation \ref{eq:const8} ensures connectivity in the proof graph, by enforcing the total flow to be $|\mathcal{N}|$. Finally, we ensure that the proof connectivity is checked on the valid edges only (which are part of the proof) through the last constraint, since a max-flow of $|\mathcal{N}|$ is achievable for any connected graph.

\section{Experiments}

Our experiments evaluate the effectiveness of \textsc{PRover} (\textsc{PR}), our joint QA, and proof model against RuleTakers (RT). Details of our experimental setup are in the appendix.

\subsection{Datasets and Evaluation Metrics}

We conduct experiments on all the three sets of datasets introduced in \citet{clark2020transformers} and consisting of gold answers and proofs. Further details of the datasets can be found in the appendix.

\noindent \textbf{DU0-DU5: }The first set consists of five datasets, each containing 100k questions with theories in synthetic language and requiring reasoning paths up to depth $D$ ($D=0,1,2,3,5$). We refer to these datasets as DU0, DU1, DU2, DU3 and DU5, where DU stands for ``Depth Upto''. 

\noindent \textbf{Birds-Electricity: }It consists of two test-only datasets of 5k samples used to evaluate the out-of-distribution performance of the models. 

\noindent \textbf{ParaRules: }
ParaRules consists of 40k questions against 2k theories expressed in paraphrased natural language, obtained through crowdsourcing.

We evaluate QA performance through accuracy. For proofs, we introduce three metrics: (1) \textbf{Node Accuracy (NA):} Fraction of examples where the predicted node set matches exactly with the gold node set, (2) \textbf{Edge Accuracy (EA):} Fraction of examples where the predicted edge set match exactly with the gold set, and (3) \textbf{Proof Accuracy (PA):} Fraction of examples where the generated proof matches exactly with the gold proof. For examples with multiple gold proofs, we give credit if the prediction matches exactly with any one of the proofs. We also evaluate \textbf{Full Accuracy (FA)}, denoting the fraction of samples where both the answer and the proof are exactly correct.

\begin{table}
\small
\centering
\begin{tabular}{lrrrrrrr}
\toprule
\multirow{2}{*}{D} & \multicolumn{1}{c}{\multirow{2}{*}{{Cnt}}} & \multicolumn{2}{c}{{QA}}                                              & \multicolumn{1}{c}{\multirow{2}{*}{{NA}}} & \multicolumn{1}{c}{\multirow{2}{*}{{EA}}} & \multicolumn{1}{c}{\multirow{2}{*}{{PA}}} & \multicolumn{1}{c}{\multirow{2}{*}{{FA}}} \\ \cmidrule{3-4}
                  & \multicolumn{1}{c}{}                                  & \multicolumn{1}{c}{{RT}} & \multicolumn{1}{c}{{\textsc{PR} }} & \multicolumn{1}{c}{}                               & \multicolumn{1}{c}{}                               & \multicolumn{1}{c}{}                                & \multicolumn{1}{c}{}                               \\ \midrule
{0}  & 6299                                                   & \textbf{100}                                 & \textbf{100}                                     & 98.6                                               & 98.5                                               & 98.4                                                &   98.4                                                  \\
{1}  & 4434                                                   & 98.4                                & \textbf{99.0}                                   & 93.3                                               & 95.1                                               & 93.2                                                &     93.1                                                \\
{2}  & 2915                                                   & 98.4                                & \textbf{98.8}                                   & 85.9                                               & 84.8                                               & 84.8                                                &    84.8                                                 \\
{3}  & 2396                                                   & 98.8                                & \textbf{99.1}                                   & 82.3                                               & 80.5                                               & 80.5                                                &    80.5                                                 \\
{4}  & 2134                                                   & \textbf{99.2}                                & 98.8                                   & 77.7                                               & 72.5                                               & 72.5                                                &   72.4                                                  \\
{5}  & 2003                                                   & \textbf{99.8}                                & 99.3                                   & 76.0                                               & 65.1                                               & 65.1                                                &   65.1                                                  \\ \midrule
{All}  & 20192                                                  & 99.2                                & \textbf{99.3}                                   & 89.2                                               & 87.5                                               & 87.1                                                & 87.1                                               \\ \bottomrule
\end{tabular}
\caption{QA comparison between RT and \textsc{PR} for varying depths along with node, edge, proof and full accuracy for \textsc{PRover} on DU5. Cnt = Sample Count.
\vspace{-10pt}}
\label{tab:depth-5}
\end{table}

\subsection{QA and Proof Results for Varying Depths}
\label{expt_1}

We first train and evaluate \textsc{PRover} on the train and test splits of the DU5 dataset, and compare its QA performance with RuleTakers for questions of varying depths (D). Table \ref{tab:depth-5} shows these results and the proof-related metrics for \textsc{PRover}. The corresponding validation set results can be found in the appendix. Overall, and at each depth, \textsc{PRover} matches the QA performance of RuleTakers. \textsc{PRover} is also able to generate exact proofs fairly accurately at 87\%. Perhaps unsurprisingly, we find that edge prediction is a harder task than node prediction, and performance worsens with increasing depth due to an increasingly large number of edges to be labeled. The proof accuracy matches the edge accuracy at each depth, suggesting that proofs are almost always correct if the edges are correct. Similarly, the full accuracy matches the proof accuracy, showing that the predicted answer is almost always correct when the corresponding proof is correct. This points to an interesting observation -- QA is easier than node prediction, which in turn is easier than edge prediction. All the datasets experimented with exhibit this behavior, as we also describe later. Proof generation becomes harder with increasing depth (and hence, more nodes and edges), as the exact proof generation accuracy drops to 65\% for depth 5. On analyzing further, we find that on average, \textsc{PRover} correctly predicts 6 out of 7 edges present in a depth 5 proof. Overall, \textsc{PRover} is interpretable yet efficient, as it generates proofs fairly accurately without any loss in QA performance.

\subsection{Zero-Shot Evaluation}
Following previous work \cite{clark2020transformers}, we now test the out-of-distribution performance of \textsc{PRover} on the Birds-Electricity dataset (Table \ref{tab:zero-shot}). The DU5-trained model is tested on six datasets, two from the birds domain (B1, B2) and another four from the electricity domain (E1, E2, E3, E4). Overall, our model achieves a 6\% QA improvement over RuleTakers. More importantly, \textsc{PRover} outperforms RuleTakers by 8\% on the hardest and largest E4 subset of the data. The proof accuracy is also fairly high, demonstrating good proof generation ability of our model for out-of-distribution data as well. Similar to the test results on DU5, the full accuracy matches the proof accuracy, demonstrating proof consistency with the predicted answers. We show examples of proofs generated by \textsc{PRover} in Figure \ref{fig:example_dataset_proof} and in the appendix.

\begin{table}[t]
\small
\centering
\begin{tabular}{lrrrrrrr}
\toprule
\multirow{2}{*}{} & \multicolumn{1}{c}{\multirow{2}{*}{Cnt}} & \multicolumn{2}{c}{QA}                                              & \multicolumn{1}{c}{\multirow{2}{*}{NA}} & \multicolumn{1}{c}{\multirow{2}{*}{EA}} & \multicolumn{1}{c}{\multirow{2}{*}{PA}} & \multicolumn{1}{c}{\multirow{2}{*}{FA}} \\ \cmidrule{3-4}
                  & \multicolumn{1}{c}{}                                  & \multicolumn{1}{c}{RT} & \multicolumn{1}{c}{\textsc{PR} } & \multicolumn{1}{c}{}                               & \multicolumn{1}{c}{}                               & \multicolumn{1}{c}{}                                & \multicolumn{1}{c}{}                               \\ \midrule 
B1       & 40 & \textbf{97.5}                                &       95.0                                                               &      92.5                                                              &        92.5                                                            &          92.5   &   92.5                         \\ 
B2 & 40      & \textbf{100}                                &     95.0                                     &             95.0                            &     95.0                                    &                95.0  & 95.0                              \\ 
E1 & 162 & 96.9                                &     \textbf{100}                                    &                                 95.1                                    &          96.3                                   &           95.1     &  95.1                            \\ 
E2 & 180 & 98.3                                &                 \textbf{100}                        &        91.7                                      &        93.3                                   &                      91.7    &    91.7              \\ 
E3 & 624 & \textbf{91.8}                                & 89.7                                        &              72.3                                    &         73.1                                    &               72.3       &  71.8                           \\ 
E4 & 4224 & 76.7                                &        \textbf{84.8}                                 &            81.4                                   &               81.3                                 &               80.6       &   80.6                  \\ \midrule
All  & 5270   & 80.1                                &    \textbf{86.5}                                     &    81.3                                     &          81.4                                   &     80.7 &   80.5                                  \\ \bottomrule
\end{tabular}
\caption{Zero-shot performance comparison on the Birds-Electricity dataset after training on DU5.}
    \label{tab:zero-shot}
\end{table}

\begin{figure}
    \centering
    \includegraphics[width=0.8\columnwidth]{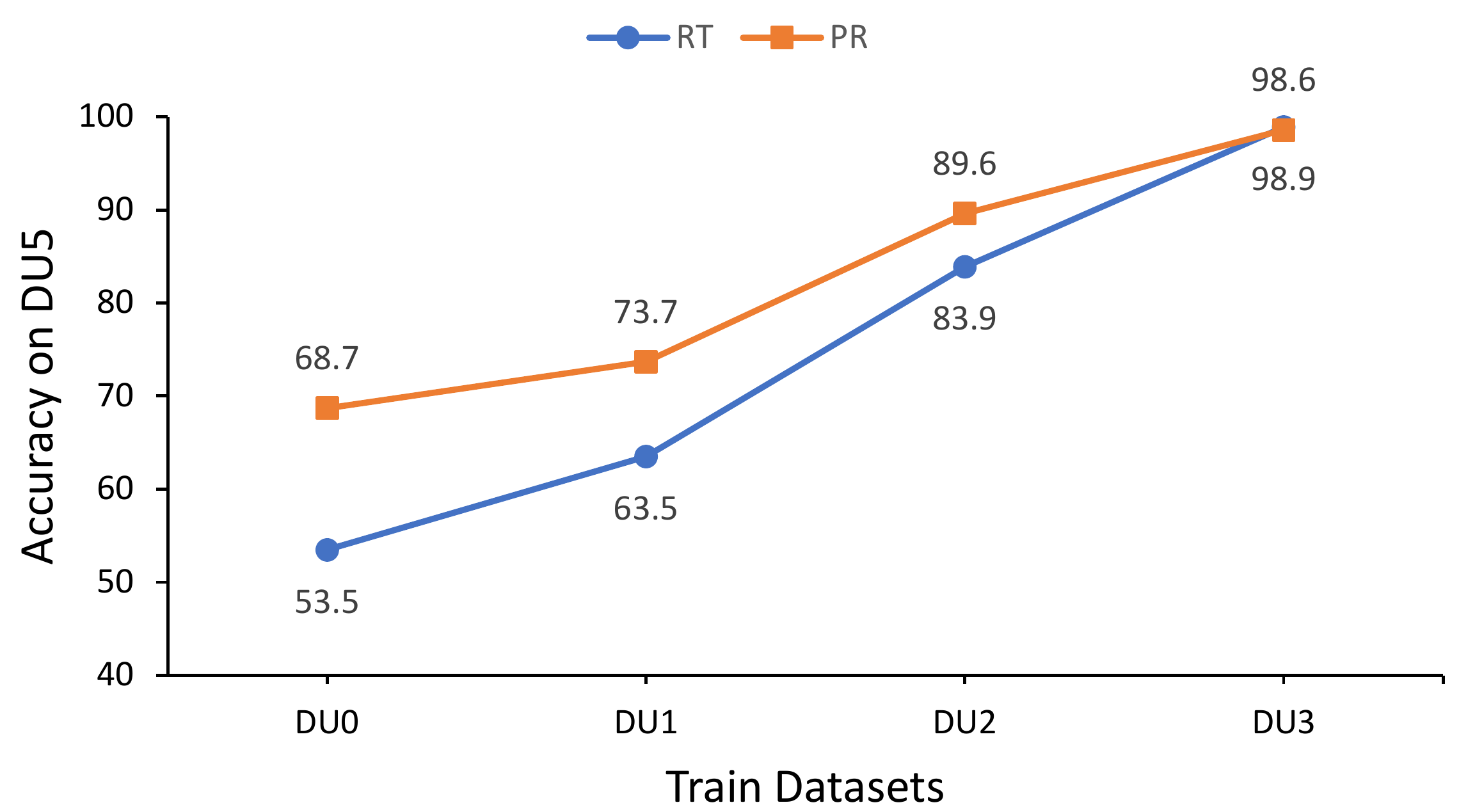}
    \vspace{-5pt}
    \caption{QA performance comparison between \textsc{PRover} and RuleTakers with models trained on DU0, DU1, DU2 and DU3 and tested on DU5.
    }
    \label{fig:gen}
    \vspace{-12pt}
\end{figure}

\subsection{Generalization to Higher Depths}

We evaluate the generalization ability of \textsc{PRover} compared to RuleTakers by training models on the train splits of DU0, DU1, DU2 and DU3, and testing the QA performance on the overall test set for DU5, which includes questions with higher depth than seen during training. The corresponding validation set and proof-related results can be found in the appendix. As shown in Figure \ref{fig:gen}, \textsc{PRover}, when trained on depth 0 examples only, performs significantly better than RuleTakers with an improvement of 15\%. A similar trend is observed for DU1 and DU2, where \textsc{PRover} improves by 10\% and 6\%, respectively. On DU3, both models show high and comparable performance. \textsc{PRover}'s superior generalization ability can be attributed to the extra training supervision incorporated in the form of proofs and an inductive bias for making proof-based predictions. While proof construction for supervised training is expensive, \textsc{PRover}'s superior QA results on out-of-distribution data (Table \ref{tab:zero-shot}) and higher depth questions is a potential first step to showing that limited proof supervision can still lead to effective generalization.

\begin{table}
\small
\centering
\begin{tabular}{lrrrrrrr}
\toprule
\multirow{2}{*}{D} & \multicolumn{1}{c}{\multirow{2}{*}{{Cnt}}} & \multicolumn{2}{c}{{QA}}                                              & \multicolumn{1}{c}{\multirow{2}{*}{{NA}}} & \multicolumn{1}{c}{\multirow{2}{*}{{EA}}} & \multicolumn{1}{c}{\multirow{2}{*}{{PA}}} & \multicolumn{1}{c}{\multirow{2}{*}{{FA}}} \\ \cmidrule{3-4}
                  & \multicolumn{1}{c}{}                                  & \multicolumn{1}{c}{{RT}} & \multicolumn{1}{c}{{\textsc{PR} }} & \multicolumn{1}{c}{}                               & \multicolumn{1}{c}{}                               & \multicolumn{1}{c}{}                                & \multicolumn{1}{c}{}                              \\ \midrule
{0}  & 2968 & \textbf{99.8} & 99.7 & 99.5 & 99.9 & 99.5 & 99.4                                               \\ 
{1}  & 2406 & \textbf{99.3} & 98.6 & 98.0 & 98.9 & 98.0 & 97.3                                         \\ 
{2}  & 1443 & \textbf{98.2} & 98.2 & 89.2 & 88.9 & 88.9 & 88.7                                               \\ 
{3}  & 1036 & \textbf{96.7} & 96.5 & 92.1 & 90.0 & 90.0 & 89.9                                                \\ 
{4}  & 142 & \textbf{90.1} & 88.0 & 87.3 & 76.1 & 76.1 & 76.1                                               \\ \midrule
{All}  & 8008 & \textbf{98.8} & 98.4 & 96.0 & 95.8 & 95.4 & 95.1                                         \\ \bottomrule
\end{tabular}
\vspace{-5pt}
\caption{Comparison of models trained on DU3 and ParaRules training sets and tested on ParaRules test set.}
\vspace{-10pt}
\label{tab:natlang}
\end{table}

\subsection{Varying Training Data Size}

We explore varying the amount of training data from 10k to 30k to all the examples (70k) in DU5. As shown in Table \ref{tab:less_data}, when trained with only 40\% of the data, \textsc{PRover} obtains a near-perfect QA accuracy of 97.8\%. Thus, for QA, \textsc{PRover}'s joint training with proofs can compensate for the lack of training data. Proof generation, however, is much harder and with increased training data, the rate of increase in proof accuracy is much more gradual.

\subsection{Evaluation on Complex Language}
We also test \textsc{PRover}'s ability to generate proofs for more human-like natural language theories. More details on the ParaRules dataset can be found in the appendix. Following \citet{clark2020transformers}, we train a model by combining the DU3 and ParaRules training partitions and test on the ParaRules test partition.
Table \ref{tab:natlang} again shows that \textsc{PRover} matches the QA performance of RuleTakers, and also generates proofs with a high accuracy of 95\%. Following previous trends, the proof accuracy drops as the depth increases, and QA performance is higher than for node prediction, which in turn is higher than for edge prediction.

\begin{table}[t]
    \small
    \centering
    \begin{tabular}{lrrrrr}
    \toprule
          \multicolumn{1}{c}{Count} & \multicolumn{1}{c}{QA} & \multicolumn{1}{c}{NA} & \multicolumn{1}{c}{EA} & \multicolumn{1}{c}{PA} & \multicolumn{1}{c}{FA} \\ \midrule
    10k & 87.1 & 48.1 & 44.7 & 44.0 & 42.7  \\ 
    30k & 97.8 & 77.9 & 73.2 & 72.5 & 72.4 \\ 
    70k & 99.3 & 89.2 & 87.5 & 87.1 & 87.1 \\ \bottomrule
    \end{tabular}
    \vspace{-5pt}
    \caption{Comparison of \textsc{PRover} models trained with varying amount of training data on DU5. Count = Number of training examples.}
    \label{tab:less_data}
    \vspace{-10pt}
\end{table}

\subsection{Ablation and Error Analysis}
\label{ablate}
Table \ref{tab:ablation} analyzes the effectiveness of the individual components of \textsc{PRover} through an ablation study. These ablated variants also provide natural baselines for our proof-related results. Specifically, we train and test the following models on DU5: (1) \textbf{QA+Node:} We train a model consisting of only the QA and Node modules; (2) \textbf{No NAF:} We train a model using random NAF embeddings; (3) \textbf{Unconstrained Train (UT) + No ILP:} We remove constraints both during training and inference; (4) \textbf{Unconstrained Train (UT) + ILP:} We remove constraints only during training; 
(5) \textbf{No Connectivity:} Finally, we train a model where we only remove the connectivity constraint during inference. More details about these models in appendix.

The QA accuracy is mostly unaffected in all our models and all but ``No NAF'' have similar node accuracy. The ``No NAF'' model does not learn a representation for NAF, leading to 5-6\% drop in both node and edge accuracy. The 5-6\% drop in edge and proof accuracy for the ``Unconstrained Train + No ILP" model, compared to \textsc{PRover}, shows that removing constraints results in a harder learning problem and the model fails to automatically learn all the constraints. The proof accuracy improves slightly when we add constraints only during inference (``Unconstrained Train + ILP''). The connectivity constraint provides only marginal improvement as our model mostly predicts connected proofs without any explicit supervision. Specifically, only 57 examples have disconnected proofs without this constraint. The overall \textsc{PRover} model outperforms all variants in full accuracy. 

To better understand the loss of accuracy for higher depth proofs, we perform error analysis of \textsc{PRover} for the depth 5 subset of DU5. We find that our NAF learning module is highly accurate -- \textsc{PRover} correctly predicts NAF in a proof 95\% of the time. Among all examples with incorrectly predicted node sets, 42\% are such that the predicted set is a subset of the gold set while for 25\% examples, it is a superset, demonstrating that our model tends to underestimate the number of essential rules and facts. \textsc{PRover} almost perfectly identifies the direction of edges. We find only 1 example where the proof is incorrect solely due to the incorrect identification of directionality. Further, 21\% of the incorrectly predicted edges are subsets of the gold sets, while 35\% are supersets.

\begin{table}[t]
    \small
    \centering
    \begin{tabular}{lrrrrr}
    \toprule
          & \multicolumn{1}{c}{QA} & \multicolumn{1}{c}{NA} & \multicolumn{1}{c}{EA} & \multicolumn{1}{c}{PA} & \multicolumn{1}{c}{FA} \\ \midrule
    {\textsc{QA+N+E (PR)}} & 99.3 & 89.2 & 87.5 & \textbf{87.1} & \textbf{87.1}  \\ 
    {QA+N} & 99.4 & 88.9 & - & - & - \\ 
    QA (RT) & 99.2 & - & - & - & - \\ 
    {No NAF} & \textbf{99.5} & 83.1 & 82.3 & 81.7 & 81.7 \\ 
    {UT + No ILP} & 99.4 & \textbf{90.1} & 83.0 & 81.9 & 81.9\\ 
    {UT + ILP} & 99.4 & \textbf{90.1} & 83.4 & 82.9 & 82.8 \\ 
    {No Connectivity} & 99.3 & 89.2 & \textbf{87.8} & 87.0 & 87.0 \\ \bottomrule
    \end{tabular}
    \vspace{-5pt}
    \caption{Ablation studies of \textsc{PRover} showing the importance of each component and constraints.}
    \label{tab:ablation}
\end{table}

\section{Discussion and Future Work}

\paragraph{Graph-based Explanations:}

While we have presented \textsc{PRover} as a model that can emulate formal reasoning, it has further potential use as an explanation generation system. \textsc{PRover} generates compositional explanations in the form of graphs and QA systems, in general, can potentially benefit from generating such graphical explanations. For example, in multi-hop QA tasks, the node module can choose all the relevant sentences in the context and the edge module can identify the flow of information between these to arrive at the answer (in the presence of task-specific constraints). Graphical explanations, in contrast to natural language ones, are more structured and can allow explicit modeling of causality (and are easier to evaluate, as opposed to free-form natural language generation). We hope that \textsc{PRover} will encourage further work towards developing interpretable NLP models with structured explanations.

\paragraph{QA and Proof Consistency:}
Currently, \textsc{PRover} predicts the answer and generates the proof by jointly optimizing the QA, node and edge modules using a shared RoBERTa model. Another modeling choice could explicitly condition the QA module on the node and edge modules so that the answer is predicted from the proof.
We empirically verify the consistency between the predicted answer and the generated proof by showing that the full accuracy matches the proof accuracy. However, in scenarios where questions have open-ended answers, generating answer from a `proof' in a consistent manner needs more exploration. \textsc{PRover}'s constraints like ensuring connectivity are necessary constraints for generating valid proofs for any graph-based explanation generation system. However, other tasks may require imposing additional constraints to ensure valid explanations.\textsc{PRover}'s inference mechanism can be extended to incorporate these.

\paragraph{Broader Implications in Formal Logic:}
\textsc{PRover}’s framework is not conceptually constrained to a particular logic fragment. \textsc{PRover} uses the idea that applying a rule to fact(s) can produce new fact(s). All logic fragments from formal logic fit this idea and may only differ in the nature of the graphs generated. For a fact ``Robin is a bird" and a rule with universal quantification ``All birds can fly", \textsc{PRover}’s graph will have an edge from the fact to the rule to generate ``Robin can fly". We experiment with datasets which already contain negations in facts. While these datasets currently do not contain disjunctions, our graphical representations of proofs allow an easy extension in such scenarios. E.g., if there is a disjunction rule ``If X or Y then Z" instead of a conjunction rule ``If X and Y then Z", only the shape of the graph changes. In the former, Z is proved by either an edge from X or from Y to the rule, while in the latter, both edges have to be necessarily present. Inferences over modals like ``might" and disjunction rules like ``If X then Y or Z" will mean that both the answer and the proof will be probabilistic. In such scenarios, \textsc{PRover}’s unweighted proof graphs can be extended to weighted ones to represent this probabilistic nature.

\section{Conclusion}
We introduce \textsc{PRover}, an interpretable joint model that answers binary questions over natural language rule-bases and generates corresponding proofs. The proofs are generated through the node and edge modules of the model in the presence of multiple global constraints during training and ILP inference. Our model improves state-of-the-art QA accuracy in the zero-shot scenario by 6\% and generates proofs accurately. \textsc{PRover} also generalizes much better to higher depth questions with up to 15\% absolute improvement in QA performance over RuleTakers. \textsc{PRover}'s modeling is relatively generic, and similar proof generation methods can be explored in traditional multi-hop QA tasks. \textsc{PRover} can also be a helpful aid to formal reasoners in scenarios where rules are fuzzy and creating rule-bases in a formal language is tedious or infeasible.

\section{Acknowledgements}
We thank the reviewers for their helpful feedback. This work was supported by DARPA MCS Grant N66001-19-2-4031, NSF-CAREER Award 1846185, DARPA YFA17-D17AP00022, ONR Grant N00014-18-1-2871, Microsoft Investigator Fellowship, and Munroe \& Rebecca Cobey Fellowship. The views in this article are those of the authors and not the funding agency.

\bibliography{emnlp2020}
\bibliographystyle{acl_natbib}

\appendix
\section{Appendix}

\subsection{Experimental Setup}

We build our model on top of the Hugging Face Transformers library \cite{wolf2019transformers}.\footnote{\url{https://github.com/huggingface/transformers}} All hyperparamters are chosen based on the best validation set performance (Full Accuracy) of the corresponding dataset. We use RoBERTa-large \cite{liu2019roberta} as the pre-trained Language Model and all our models are trained using a batch size of $8$ and a maximum sequence length of $300$. We train the models for a maximum of 5 epochs using an initial learning rate of $10^{-5}$, with linear decay and a weight decay of $0.1$ . The dropout probability is chosen to be $0.1$. The random seed used in all the experiments is $42$. Each epoch of \textsc{PRover} takes $2.5$ hours to run on one V100 Volta GPU. The total number of parameters of \textsc{PRover} is similar to that of RoBERTa-large ($355M$). Batch size and learning rate are manually tuned in the range \{$8$,$16$\} and \{$10^{-5}$, $2*10^{-5}$\} respectively.
The ILP is modeled using PuLP.\footnote{\url{https://pypi.org/project/PuLP/}} Proofs in the datasets are represented as bracketed strings, which are pre-processed into graph representations consisting of unique nodes and edges. The maximum number of facts and rules corresponding to a context is $25$.\footnote{Further details of our best hyperparameters can be found in the attached code as part of the supplementary material.}

\begin{table*}
\centering
\small
\begin{tabular}{lrrrrrrrrrr}
\toprule 
              & \multicolumn{2}{c}{{QA}}                                       & \multicolumn{2}{c}{{NA}}                                     & \multicolumn{2}{c}{{EA}}                                     & \multicolumn{2}{c}{{PA}}                                    & \multicolumn{2}{c}{{FA}}                                     \\ \midrule 
             & \multicolumn{1}{c}{{Dev}} & \multicolumn{1}{c}{{Test}} & \multicolumn{1}{c}{{Dev}} & \multicolumn{1}{c}{{Test}} & \multicolumn{1}{c}{{Dev}} & \multicolumn{1}{c}{{Test}} & \multicolumn{1}{c}{{Dev}} & \multicolumn{1}{c}{{Test}} & \multicolumn{1}{c}{{Dev}} & \multicolumn{1}{c}{{Test}} \\ \midrule
{DU0} &     68.3                              &         68.7                           & 45.3                                  &    46.0                                &   49.3                                &      49.5                              &  43.8                                 &    44.4                                &            42.3                       &      42.8                                                     \\ 
{DU1} &    73.2                               & 73.7                                    & 66.4                                  &  66.3                                  &  64.5                                 &  64.3                                  &    63.9                               &   63.8                                 & 61.8                                  &       61.9                                                   \\ 
{DU2} & 89.3                                  & 89.6                       &   76.6          &  76.4                                 & 73.1                                   &     73.1                              &    72.6                                & 72.6                                  & 72.3                                   &  72.3                                                                                          \\ 
{DU3} &    98.3                               & 98.6                   &   85.5              &   85.0                                & 79.9                                   &  79.5                                 &    79.4                                &  79.1                                 & 79.4                                   &    79.1                               \\ \bottomrule
\end{tabular}
\caption{ Performance of \model{} trained on the training splits of DU0, DU1, DU2 and DU3 and tested on the validation  and test splits of DU5.}
    \label{tab:gen_all}
\end{table*}

\begin{table}
\small
    \centering
    \begin{tabular}{lrrrrrr}
    \toprule
         \multicolumn{1}{l}{D} & \multicolumn{1}{c}{Cnt} & \multicolumn{1}{c}{QA} & \multicolumn{1}{c}{NA} & \multicolumn{1}{c}{EA} & \multicolumn{1}{c}{PA} & \multicolumn{1}{c}{FA} \\  \midrule
         {0} & 3116 & 100 & 98.7 & 98.6 & 98.5 & 98.5 \\ 
         {1} & 2304 & 98.8 & 92.5 & 94.9 & 92.2 & 92.2 \\ 
         {2} & 1436 & 99.2 & 86.1 & 85.6 & 85.6 & 85.6 \\ 
         {3} & 1165 & 98.7 & 85.1 & 82.8 & 82.8 & 82.8 \\ 
         {4} & 1041 & 98.8 & 81.2 & 76.9 & 76.9 & 76.9 \\ 
         {5} & 990 & 99.3 & 78.3 & 67.4 & 67.4 & 67.4 \\ \midrule
         {All} & 10068 & 99.3 & 90.0 & 88.6 & 88.0 & 88.0 \\ \bottomrule
    \end{tabular}
    \caption{ Performance of \model{} trained on the training split of DU5 and tested on the validation split of DU5.}
    \label{tab:dev_du5}
\end{table}

\subsection{Dataset Details}
Below we briefly describe the three sets of datasets we conduct experiments on.\footnote{\url{ https://rule-reasoning.apps.allenai.org/}} Each dataset has a train, validation and test split, except for the zero-shot test-only one. Further details about these can be found in \citet{clark2020transformers}.

\paragraph{\textbf{DU0-DU5: }}The first set consists of five datasets, each containing 100k questions with theories in synthetic language and requiring reasoning paths up to depth $D$ ($D=0,1,2,3,5$). For example, $D=0$ means the true facts can be proved by simple lookup in the context. The samples are randomly split 70/10/20 into train/dev/test partitions such that there is no overlap of theories between the partitions.

\paragraph{\textbf{Birds-Electricity: }}The second set consists of two test-only datasets used to evaluate robustness and out-of-distribution performance of the models. The contexts are about birds and an electric circuit, and consist of 5k samples in total. The vocabulary of entities, attributes and predicates, apart from \texttt{is()} are all new at test time.

\paragraph{\textbf{ParaRules: }}
The final dataset, ParaRules consists of 40k questions against 2k theories expressed in paraphrased natural language, obtained through crowdsourcing. While the previous datasets contain synthetic language, ParaRules tests the models' ability to reason over more human-like paraphrased language.

\subsection{QA and Proof Results for Varying Depths}

Table \ref{tab:dev_du5} shows the DU5 validation set performance of \model{} trained on the training split of DU5. \model{} obtains a near perfect QA accuracy and a proof accuracy of 88\%. While the QA accuracy remains equally high at all depths, the proof accuracy drops with increasing depth. Full accuracy matches the proof accuracy, demonstrating consistency between the predicted answers and generated proofs.

\subsection{Generalization to Higher Depths}

In Table \ref{tab:gen_all}, we provide detailed results of \textsc{PRover}'s generalization ability to higher depth questions. Specifically, we evaluate four models, trained on the training splits of DU0, DU1, DU2 and DU3 and tested on the validation and test splits of DU5. We have shown previously that \textsc{PRover} does significantly better than RuleTakers \cite{clark2020transformers} on QA generalization. The proofs, however, do not generalize that well. Note that depth 0 proofs are rather simple (consisting of a single fact) and a model trained on only such proofs, unsuprisingly, fails to generate proofs for higher depth questions. However, the proof results start improving as the model gets trained on more complex proofs and reaches an accuracy of 79\%, when trained on DU3 questions.

\begin{table}
\small
    \centering
    \begin{tabular}{lrrrrrr}
    \toprule
         \multicolumn{1}{l}{D} & \multicolumn{1}{c}{Cnt} & \multicolumn{1}{c}{QA} & \multicolumn{1}{c}{NA} & \multicolumn{1}{c}{EA} & \multicolumn{1}{c}{PA} & \multicolumn{1}{c}{FA} \\ \midrule
         {0} & 1485 & 99.9 & 99.6 & 99.7 & 99.5 & 99.5 \\ 
         {1} & 1180 & 99.7 & 99.3 & 99.5 & 99.3 & 99.3  \\ 
         {2} & 727 & 99.4 & 91.5 & 91.5 & 91.5 & 91.3  \\ 
         {3} & 524 & 98.5 & 92.0 & 90.3 & 90.3 & 90.3  \\ 
         {4} & 81 & 100 & 87.6 & 72.8 & 72.8 & 72.8 \\ 
         {5} & 7 & 100 & 100 & 0 & 0 & 0   \\ \midrule
         {All} & 4004 & 99.6 & 96.8 & 96.2 & 96.1 & 96.0 \\ \bottomrule
    \end{tabular}
    \caption{\model{} results on the ParaRules validation set after training on DU3+ParaRules training splits.}
    \label{tab:dev_natlang}
\end{table}

\subsection{Evaluation on Complex Language}

In Table \ref{tab:dev_natlang}, we report the ParaRules validation set results of \textsc{PRover} trained on the combination of DU3 and ParaRules training splits (following previous work \cite{clark2020transformers}). ParaRules is created by first separating the fact groups (a fact group is the set of all facts in the theory concerning a  particular person) and the rules from a theory and then asking crowdworkers to paraphrase these in their own words. For example, a fact group ``Alan is blue. Alan is rough. Alan is young.'', may be reworded into ``Alan is on the young side, but rough. He often feels rather blue.''. Thus, unlike the previous datasets where the proof graphs are composed of facts and rules, ParaRules proofs are composed of fact groups and rules.\footnote{The original ParaRules dataset released by \citet{clark2020transformers} has proofs for the unparaphrased theories (consisting of facts and rules). Using the mapping from a fact to the corresponding fact group, we replace the fact nodes in the proof graph with the corresponding fact group nodes. Note that this is done to report proof accuracy for this dataset as well.} \model{} obtains high QA and proof accuracy on complex human-parapharsed rule-bases, showing good generalization on such language. However, the proof accuracy again drops as the depth of the questions increases.

\subsection{Ablation Models and Simpler Baselines}

We provide brief descriptions of our ablation models. These are (1) \textbf{QA+Node:} We model \textsc{PRover} consisting of only the QA and Node modules. Since there is no edge module, this model does not require any constrained training or inference; (2) \textbf{No NAF:} We train a model using random NAF embeddings with no learning. This helps us understand the effectiveness of our NAF learning; (3) \textbf{Unconstrained Train + No ILP:} Through this model, we study the effectiveness of our global constraints. Specifically, no edges are masked for training and during inference, the edge labels are predicted based on the model's probability scores only; (4) \textbf{Unconstrained Train + ILP:} Here the constraints are employed only during inference.  Note that the reverse configuration, constrained training without ILP inference, is not included as the edge logits for the masked out labels would be random (since they are not learned). 
(5) \textbf{No Connectivity:} Finally, we train a model where we only remove the connectivity constraint during ILP optimization, keeping everything else same.

We also experiment with simpler baselines for edge prediction like training a Random Forest with lexical features (BLEU scores, length difference, word overlap, etc.) and this obtains a much lower edge accuracy of 47\%. This fails primarily because (1) proof graphs can contain NAF  which account for 9\% of the data and edges from it cannot be learned without learning a latent representation; (2) overlap features are mostly symmetric and hence are not enough for learning directionality; (3) there is lack of overall context information.

\subsection{Critical Sentence Identification}

\citet{clark2020transformers} provide an initial solution towards generating explanations for the predicted answers by using a post-hoc method -- they remove each fact or rule from the theory and check if the predicted answer changes with the new theory. They define all such rules and facts which flip the answer as critical sentences. If an example has multiple gold proofs, a critical sentence is one which is present in all the proofs. We argue that this leave-one-out analysis is not ideal for multiple reasons - (1) This does not work if the theory has negations, (2) This only predicts the presence or absence of rules and facts, and does not look at the entire chain of reasoning, which our model achieves through the edge module.
\begin{table}[h]
    \small
    \centering
    \begin{tabular}{lrrrr}
        \toprule
         \textbf{} & \multicolumn{1}{c}{Accuracy} & \multicolumn{1}{c}{Precision} & \multicolumn{1}{c}{Recall} & \multicolumn{1}{c}{F1} \\ \midrule
         {RuleTakers} & 74.5 & \textbf{98.7} & 86.9 & 92.4 \\
         {\textsc{PRover}} & \textbf{78.1} & \textbf{98.7} & \textbf{87.2} & \textbf{92.6}\\
         \bottomrule
    \end{tabular}
    \caption{Comparison of critical sentence identification on the No Negation subset of DU5 test set.}
    \label{tab:leave-one-out}
\end{table}
In our final experiment, we still apply the leave-one-out-strategy on the no-negation subset of the DU5 test set for a direct comparison with RuleTakers. As shown in Table \ref{tab:leave-one-out}, our model identifies the exact critical sentences in an example in 78\% of the cases, a 4\% improvement over RuleTakers.

\begin{figure*}[t]
\centering
    \includegraphics[width=\textwidth]{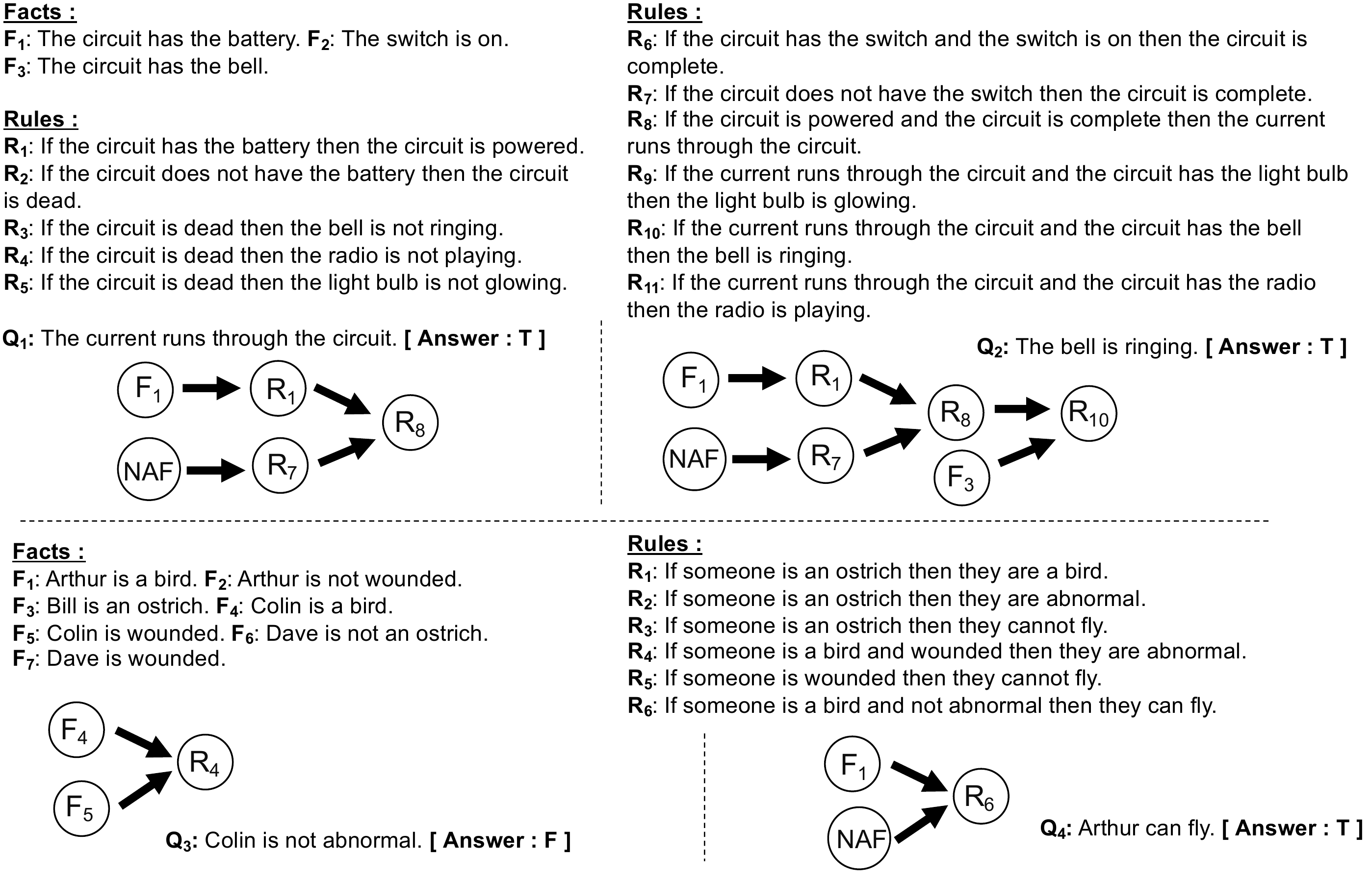}
    \vspace{-5pt}
\caption{Examples of proofs generated by \textsc{PRover} for four questions on two rule-bases about electric circuits and birds from the Birds-Electricity dataset. \textsc{PRover} not only answers the questions correctly but also accurately predicts the long reasoning chains with multiple branches.}\vspace{-5pt}
\vspace{-5pt}
\label{fig:example_bird_elec_proof}
\end{figure*}

\begin{figure*}[t]
\centering
    \includegraphics[width=\textwidth]{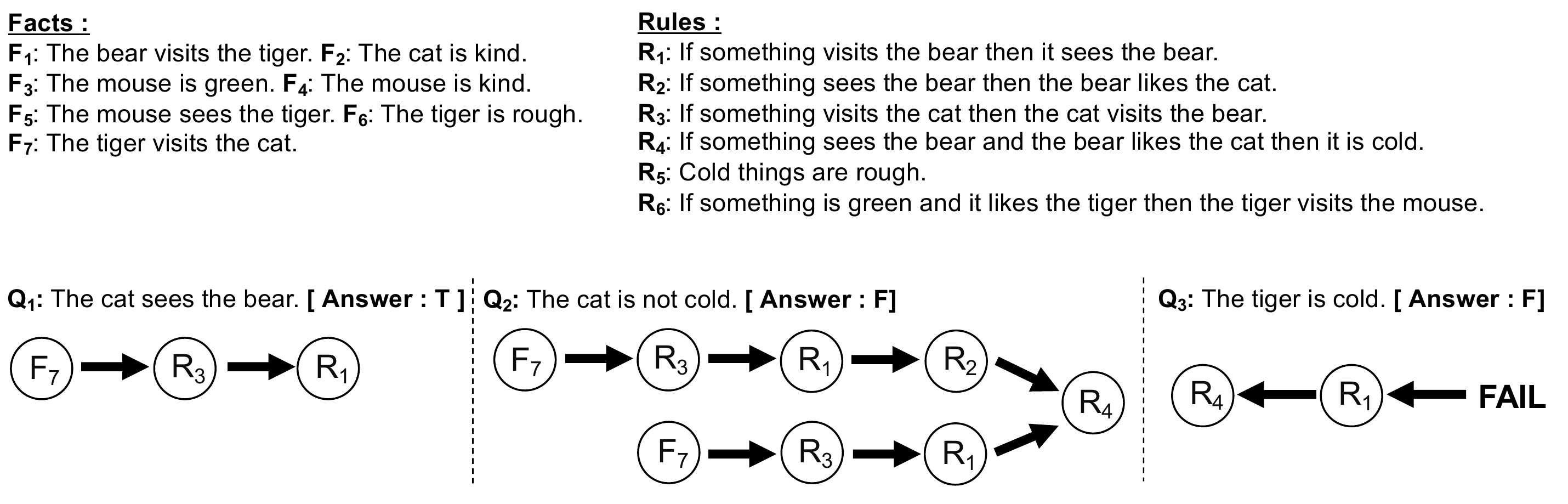}
    \vspace{-5pt}
\caption{Examples of proofs generated by \textsc{PRover} for three questions on a rule-base from the DU5 dataset. The proof corresponding to the last question is a failed case.}\vspace{-5pt}
\vspace{-5pt}
\label{fig:example_du5_proof}
\end{figure*}

\subsection{Proofs Generated by \textsc{PRover}}

In Figure \ref{fig:example_bird_elec_proof}, we show two rule-bases, one about electric circuits and another about birds from the Birds-Electricity dataset. \textsc{PRover} not only answers the questions correctly but also generates the proofs accurately. These proofs are complex because of the presence of NAF and also the long chains of reasoning needed in the inference process. Figure \ref{fig:example_du5_proof} shows three more accurate proofs generated by \textsc{PRover} for three questions from the DU5 datset.

\end{document}